\RequirePackage{rotating}
\documentclass[12pt]{extarticle}

\usepackage{booktabs} 

\usepackage[ruled]{algorithm2e} 
\usepackage{rotating}
\usepackage{csvsimple}
\usepackage[margin=1in]{geometry}
\usepackage{newtxtext}
\usepackage[utf8]{inputenc}
\usepackage[english]{babel}

\SetAlFnt{\small}
\SetAlCapFnt{\small}
\SetAlCapNameFnt{\small}
\SetAlCapHSkip{0pt}
\IncMargin{-\parindent}

\usepackage[cmex10]{amsmath}
\usepackage{multirow}
\usepackage[caption=false,font=footnotesize]{subfig}
\usepackage{csvsimple}
\usepackage{array}
\usepackage{dblfloatfix}
\usepackage{longtable}
\usepackage{supertabular}
\usepackage[hidelinks]{hyperref}
\usepackage[numbers, sort&compress]{natbib}
\usepackage{titling}

\usepackage{amssymb}
\usepackage{tabularx}
\usepackage{gensymb} 

\newcommand{\tabitem}{\par\hspace*{\labelsep}\textbullet\hspace*{\labelsep}\hangindent=6mm}

\hypersetup{
  pdfauthor={Ebenezer R.H.P. Isaac},
  pdftitle={Trait of Gait -- A Survey on Gait Biometrics},
  pdfsubject={Gait Analysis}
}

\thanksmarkseries{alph}

\title{Trait of Gait: A Survey on Gait Biometrics}

\date{\small \today}

\author{\normalsize  Ebenezer R.H.P. Isaac\thanks{\textit{Indian Institute of Technology Madras, Chennai, India.} e-mail: \texttt{ebeisaac@ieee.org}},
  \hspace{5pt}\normalsize Susan Elias\thanks{\textit{Vellore Institute of Technology, Chennai Campus, India.} e-mail: \texttt{susan.elias@vit.ac.in}},
  \hspace{5pt}\normalsize Srinivasan Rajagopalan\thanks{\textit{Mayo Clinic, Rochester, MN, USA.} e-mail: \texttt{Rajagopalan.Srinivasan@mayo.edu}},
  \hspace{2pt} and \normalsize K.S. Easwarakumar\thanks{\textit{Anna University, Chennai, India.} e-mail: \texttt{easwarakumarks@gmail.com}}
}

\providecommand{\keywords}[1]
{
  \small	
  \textbf{\textit{Keywords---}} #1
}

\begin{document}

\maketitle
  
\begin{abstract}
  Gait analysis is the study of the systematic methods that assess and quantify animal locomotion. The research on gait analysis has considerably evolved through time. It was an ancient art, and it still finds its application today in modern science and medicine. This paper describes how one's gait can be used as a biometric. It shall diversely cover salient research done within the field and explain the nuances and advances in each type of gait analysis. The prominent methods of gait recognition from the early era to the state of the art are covered. This survey also reviews the various gait datasets. The overall aim of this study is to provide a concise roadmap for anyone who wishes to do research in the field of gait biometrics.

  \vspace{1em}

\end{abstract}

\section{Introduction}
\label{sec:intro}

The simplest definition of gait states that it is the manner and style
of walking \cite{whittlegait}. This can refer to any animal that can
walk whether bipedal or quadrupedal. It can be more sophisticatedly
defined as the coordinated cyclic combination of movements that result
in locomotion \cite{boyd2005biometric}. This definition can be equally
applicable to any form of activity that is repetitive and coordinated
so as to cause motion of the living being originating it. Gait can
vary from walking, running, climbing and descending the stairs,
swimming, hopping and so on; all of which follows the same principle
in this definition. In the context of this survey, the word
`gait' generally refers to the `walking gait' of a human.

Recent research has proved that gait can be used as a non-obtrusive form of biometric \cite{boyd2005biometric}. The Defense Advanced Research Projects Agency (DARPA) launched the Human Identification at a Distance (HumanID) programme in the year 2000 (ended in 2004). This research programme focused on human identification through the face, gait and the application of new technologies. The intention of the programme was to use the state of the art unobtrusive distance biometric techniques on real world data. Gait-based recognition played a significant role in here. Major institutions which took part in this program are MIT, CMU, USF, UMD, Georgia Institute of Technology, and the University of Southampton \cite{hidgait}. The datasets compiled by most of these institutions are publicly available. Section~\ref{sec:gait-datasets} discuss more on those datasets.

\begin{figure}
  \centering 
  \includegraphics[width=0.5\linewidth]{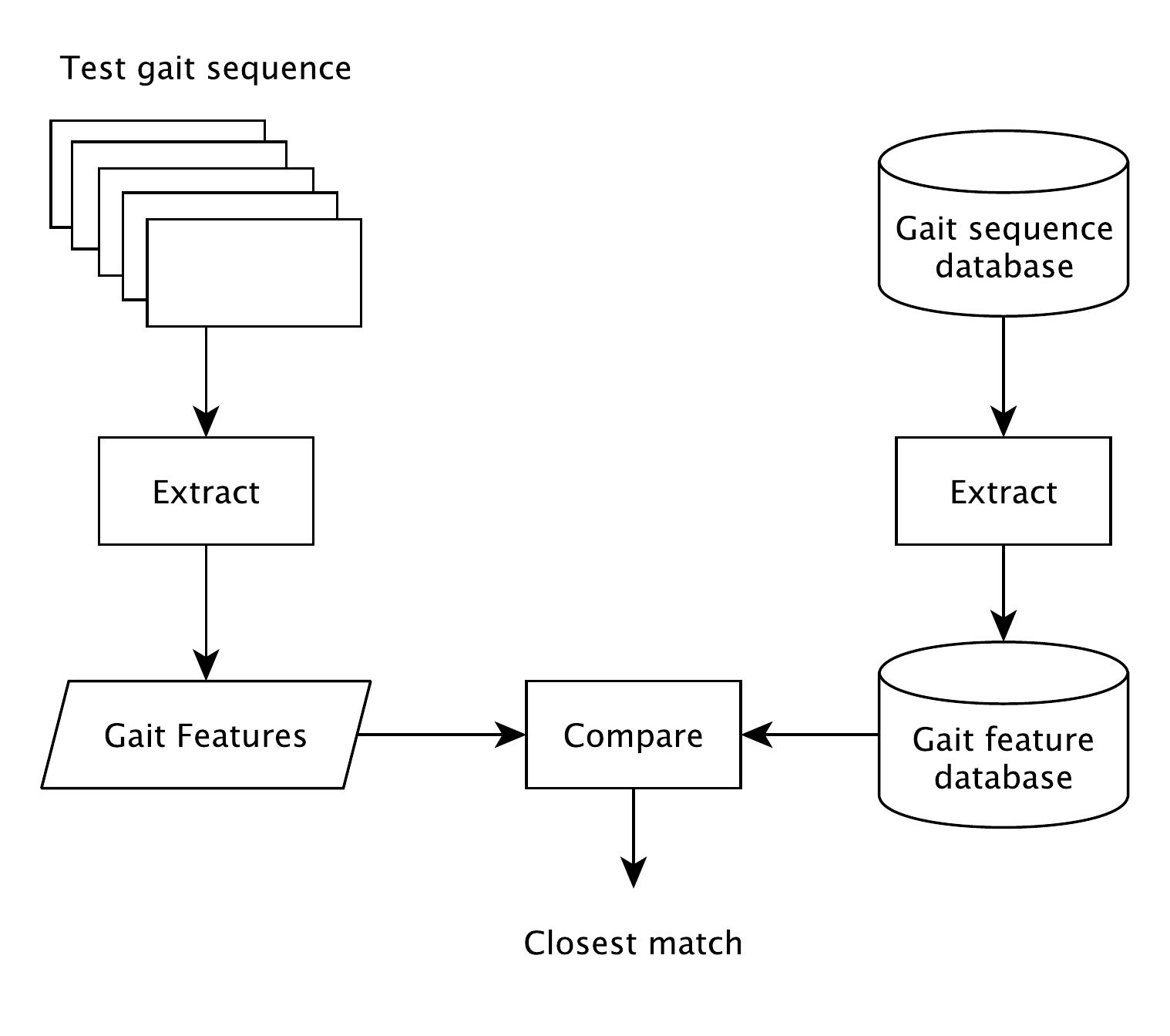}
  \caption{Simplified gait recognition system}
  \label{fig:gaitrecog}
\end{figure}

Gait is characteristic to one's individuality it is considered to be
an accepted form of unobtrusive biometrics
\cite{boyd2005biometric}. Hence, in theory, it does not require the
cooperation nor the awareness of the individual being observed. A
simplified template for a gait recognition system is as shown in
Figure~\ref{fig:gaitrecog}. A gait sequence is a temporal record of an
person's gait. It can be a sequence of frames of a video or
point-light motion, or foot pressure and foot placement pattern. These
are usually translated into a feature database after some means of
preprocessing and feature extraction. When the test gait sequence is
given for identification, the system refers the gait feature database
and returns the closest match (if any) according to some criteria. If
the similarity between test gait and stored gait are used as a measure,
the closeness can be determined by formulating a threshold. Gait
recognition systems are sometimes evaluated in terms of the ranks to
which the true identities are resolved. For instance, rank-1 implies
the true gait is identified as the first closest match, while rank-$k$
implies that the true gait was identified within the $k$ closest
matches from the database \cite{man2006individual}.

Gait recognition can be broadly classified into two types: model-based
and model-free. Due to its simplicity and efficiency in use,
model-free methods are more prevalent. We further categorise
model-free techniques as template-based and non-template
methods. 

Many benchmark datasets are available to compare the performance of
one algorithm with another. The details of such databases are available in section~\ref{sec:gait-datasets}.

 
\section{Inhibiting Factors}
\label{sec:inhibiting-factors}

Certain factors inhibit the effectiveness of biometric gait
recognition by confounding the features that can be observed
\cite{boyd2005biometric}. The following briefly explains each factor
with relevant research on how they deviate the parameters of gait.

\textbf{Walking surface}. Studies show that when the surface is
unstable, such as a wet surface, the walking speed, toe-in angle and
step length are significantly reduced to retain control over balance
\cite{menant2009effects}. However, when walking on other irregular surfaces
like grass, foam or studded with small obstacles, walking speed can be
maintained with a variable cadence and a longer stride length
\cite{menz2003acceleration}. Even two regular surfaces such as vinyl
and carpet have a significant difference
\cite{RozinKleiner2015147}. In slippery surfaces, reductions can also
be observed in stance duration, load supporting foot, normalised
stride length \cite{cham2002changes}.

\textbf{Footwear}. When the footwear is considerably different, so is
the gait of the individual. This especially concerns high heel
users. To maintain their stability, people wearing high heels require
more control over their centre of mass \cite{Chien20141045} and tend
to have longer double support times \cite{menant2009effects}. Recent
studies show that habitually shod walkers and habitual barefoot walkers
exhibited a significant gait difference when switching their footwear
in terms of their stride length and cadence
\cite{franklin2015barefoot}.
 
\textbf{Injury}. When any portion of the lumbar region or lower limb
is injured, the individual naturally adopt an antalgic gait. The individual walks in a
way so as to avoid the pain caused by the injury. This style of
walking restricts the range of motion of the associated limb leading
to a deviation in the usual gait proportional to the magnitude of the
injury. In-depth research on how injury affects gait is provided by
Ruoli Wang's thesis \cite{wang2012biomechanical}.

\textbf{Muscle development}. The development of muscles gives a
different range of control over the parts of the body that affects
gait. The sheer mass of the muscles developed alter the centre of mass
at the associated mobile limb as well as the body itself. The shift
depends on the difference in mass. The change in the centre of mass
can modify the inclination of pressure required for proper stability
\cite{lee2006detection}. An extensive study on the correlation between
muscle mass with gait performance \cite{beaudart2015correlation}
proves that muscle mass directly relates to gait speed, especially in
the case of geriatric patients.
 
\textbf{Fatigue}. When the individual is subjected to fatigue, the
stability of the concerned gait decreased while a noticeable increase
in variability of gait is exhibited \cite{vieira2016effects}. The time
taken for the recovery towards normal gait depends on the extent of
exertion applied to the individual so as to get to a fatigued state
and the individual's stamina. This aspect was observed from Ashley
Putnam's thesis \cite{putnam2013effects} in which a study with army
cadet treadmill protocol was conducted to analyse the effect of
exhaustion on gait mechanics and possible injury. Cadets ran till
exhaustion and had their gaits observed. The resulting gait had both
inter- and intra-variable vertical stiffness in the lower limbs.

\textbf{Training}. When the individual is subjected to some form of
physical training, it is possible that her/his gait is also subjected
to change. This change can be evident as a result of military
training, prolonged load condition, prolonged use of particular
footwear and athletic training.

\textbf{Extrinsic control}. Humans have an ability to control their
gait to an extent so as to differ from their usual gait. A person
can mince walk, and depending on how self-aware the individual may be
she/he can walk with a swagger or strut, a brisk walk or
tip-toe. Another matter to note is the level of awareness the
individual have of his/her surroundings. This would correspond to the
tendency to alter gait while self-control correlates with the degree
to which the gait can be altered by the individual. This factor explains how
members of the army can synchronise gait during a march.

\textbf{Intrinsic control}. There are elements that can control a
person's gait in such a way that the individual is sometimes unaware
of the change that takes place. The best example of this case are
emotional responses or mood of the individual: state of happiness,
sorrow, anger or any other emotion strong enough to make an impact on
one's gait. The variability can range from subtle to significant and
can vary from one person to another. Related studies are provided in
\cite{montepare1987identification}.

\textbf{Age}. Although the factor may not contribute to change in gait
over a short period, it certainly does influence gait to a
large extent. Ageing, in general, causes musculoskeletal and
neuromuscular losses. To compensate for these losses, the
individual make certain adjustments which can be observed in the
individual's gait \cite{MonizPereira2012S229}.

\textbf{Clothing}. While the change of clothing does not necessarily
modify the gait for slight differences in weight, it might, however,
show changes in the associated silhouettes. This change would affect a
major portion of gait recognition algorithms that depends on the
spatial configuration of silhouettes. However, a greater change in the
weight of the clothing, such as a winter suit, has a higher
probability of affecting the gait itself.

\textbf{Load}. The effect of load can significantly influence gait. In
a loaded condition such as wearing a backpack, the individual is
subjected to a higher weight in addition to his/her body
weight. In order to regulate locomotion, the foot exerts higher
pressure during plantarflexion generating a greater ground reaction
force than the unloaded condition \cite{Castro201541}. Apart from the
pressure applied, the body must cope with the change in balance for a
stable gait \cite{mummolo2016computational}. The load can also be
asymmetrical, such as a wearing a handbag, cross-bag or shoulder bag,
or carrying a suitcase. In this case, a greater difference in the
pelvic rotation is observed \cite{hyung2016influence}. The body shifts
the pelvic movement so as to counteract the imbalance caused by the
load.

\section{Model-free techniques}
\label{sec:spatio}

\begin{figure*}
  \centering
  \includegraphics[width=\linewidth]{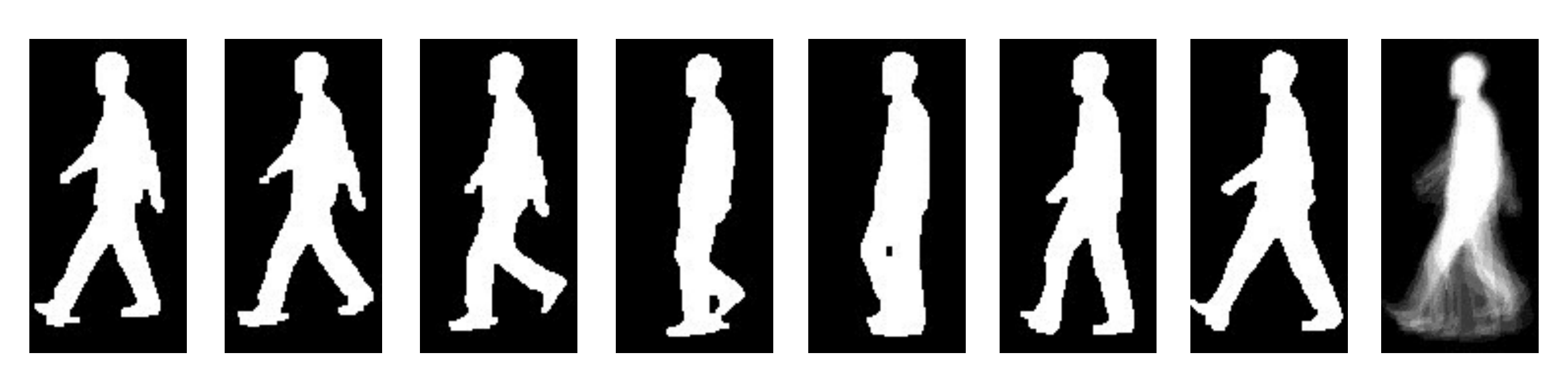}
  \caption{The Gait Energy Image of a gait sequence over a single gait
    period. The last image is the GEI template where the first seven
    sequences depict successive normalised frames of gait.}
  \label{fig:GEIseq}
\end{figure*}

The vast majority of model-free techniques tend to have a strong
reliance on spatiotemporal analysis of silhouettes of the individual
during gait. A spatiotemporal analysis takes into account the
variation in the spatial domain with respect to that in the time
domain. So when this is applied in gait recognition, the analyses
involves the observation of the spatial locations of body parts and
their movement in different stages in time.

Nearly all model-free methods have background subtraction and
silhouette extraction as the first step. Background subtraction is a
simple method in which the change in pixel values between one frame
and the successive frame is observed to bring out only the objects
that are seen in motion. From these objects, the
moving human silhouette can be extracted. The result of background
subtraction is usually binarized in which the moving object seems to
be white and the background is black, or vice versa in some cases like
the USF HumanID silhouette database.

The novelty mainly lies in how the features are extracted. The
recognition process involves the use of an established machine
learning algorithm such as the nearest neighbour classifier (used in
\cite{Hayfron_A, niyogi1994analyzing, Lam_T,Yang_X, lam2011gait,
  Lee2015, Xing2015, Arora2015}), and HMM (used in \cite{Zongyi_L,
  kale2002framework, liu2006improved, Zhang_R, Nguyen2014}). Some
apply different techniques such as canonical analysis
\cite{Huang_PS,Foster_JP}, spatiotemporal correlation
\cite{Murase_H}, and DTW
\cite{tanawongsuwan2001gait,lam2011gait,Choudhury_SD}. A few others
devise newer techniques for closeness representation.

\vspace{5pt}
\subsection{Earlier methods}
\label{sec:earlier-methods}

The earliest known spatiotemporal gait recognition techniques were
published during the 90s. Niyogi et al. \cite{niyogi1994analyzing}
proposed to recognise gait at a sagittal angle with the subject
walking frontoparallel. It is a template-based method which modelled
the human gait in the form of a set of spatiotemporal snakes
\cite{kass1988snakes} from the slices of the moving parts of the human
contour along the time domain. They expressed the spatiotemporal
dimensions as XYT signifying the 2D $(x,y)$ spatial coordinates
varying with respect to time $t$. The recognition they have obtained
with 26 image sequences across five human subjects reaches up to 83\%.

J. Little and J. Boyd \cite{little1998} introduced the concept of
optical flow in gait recognition. In principle, the points in the
image sequence that vary with time tend to oscillate periodically
during the subject's gait. By observing the optical flow of these
points, the $m$ time varying scalars can be produced. From these
scalars, the phases of the oscillations $\phi_1, \phi_2, ..., \phi_m$
can be extracted. The phases are normalised with respect to a
reference phase $\phi_m$ to form the feature vector
$F = (F_1, F_2, ..., F_{m-1})$ where $F_i = \phi_i - \phi_m$. Hence
$F$ is used to represent the gait instance. They reached a higher
recognition rate of up to 95\% by testing their technique with seven
instances for each of six human subjects.

An earlier version of template matching method was proposed by
P. S. Huang et al. \cite{Huang_PS}. They used the Eigenspace
transformation (EST), as adopted by Murase and Sakai \cite{Murase_H},
to convert the gait taken as a sequence of images to a template called
the `eigengait'. On top of this, canonical space transformation is
applied with generalised Fisher linear discriminant function to
separate the classes boundaries required for prediction. However, they
also seemed to use the small dataset used by Little and Boyd for their
application to show a questionable accuracy of 100\%.

Static features are those that do not change over the temporal
domain. These include features such as the height of the person, the
length of the limbs. Johnson and Bobick \cite{johnson2001multi} used
static features to obtain a recognition rate of 94\% with an expected
confusion of 6.37\%. The experiment involved placing cameras at each
of two different locations; near and far. 18 human subjects were made
to walk three times each and were observed at the two oblique angles.
The features determined were the height of the bounding box, the Euclidean
distance between the head and pelvis, maximum length between the
pelvis and either of the feet, and the distance between the left and
right foot.
 
Though the above methods show attractive recognition rates, all the
methods proposed at those times before suffered one major drawback:
their accuracies are biased to their samples which are too small
when considering the application as a biometric.

\vspace{5pt}
\subsection{Template-based methods}
\label{sec:gei-based}

These are a class of methods that involves transforming the sequence
of silhouette images taken from a gait video to a single image that
holds the composition of the motion-related features of the sequence.

Hayfron-Acquah et al. \cite{Hayfron_A} assessed the symmetry of the
extracted silhouette using a generalised symmetry operator. The
contours of the silhouette are extracted by applying Sobel edge
detection function. From the sequence of contours obtained, a symmetry
map is produced. Euclidean distance between Fourier descriptors is
used as a similarity measure for gait recognition. They used the SOTON
database for their experimentation and had attained a correct
classification rate (CCR) of 97.3 \% for $k=1$ and 96.4\% $k=3$ using
the nearest neighbour classifier.

\begin{sidewaystable}
  \centering
  \caption{Template-based Methods for Gait Recognition}
  \label{tab:template-based}
  \begin{minipage}{0.75\columnwidth}
    \footnotesize
    \begin{center}
      \begin{tabular}{rcllll}
        \toprule S.No. & Year & Study & Technique & Plane & Dataset \\
        \midrule
        \begin{filecontents*}{GR-T.csv}
Study,Year,Type,Technique,Plane,Dataset
Niyogi et al. \nocite{niyogi1994analyzing},1994,TS,Fitting of spatiotemporal snakes,Sagittal,$5s \sim 26$ seq.
Murase \& Sakai \nocite{Murase_H},1996,TS,Parametric eigenspace transformation,Sagittal,$7s \times 10i = 70$ seq.
P. Huang et al. \nocite{Huang_PS},1999,TS,Canonical and eigenspace transformation,Sagittal,UCSD
L. Wang et al. \nocite{wang2003automatic},2003,TS,Procrustes shape analysis,Multiview,CASIA-A
S. Sarkar et al. \nocite{sarkar2005humanid},2005,TS,Gait Baseline,Sagittal,{USF, CMU}
Ju Han \& Bir \nocite{man2006individual},2006,TS,Gait Energy Image (GEI),Sagittal,{USF v1.7, v2.1}
T. Lam et al. \nocite{Lam_T},2007,TS,MSCT and SST,Sagittal,{USF v1.7, v2.1, SOTON}
X. Yang et al. \nocite{Yang_X},2008,TS,Enhancing GEI through dynamic regions,Sagittal,USF v2.1
Kusakunniran et al. \nocite{Worapan2009auto},2009,TS,Weighted Binary Pattern (WBP),Sagittal,{CMU, CASIA-A,C}
Bashir et al. \nocite{bashir2010gait},2010,TS,Gait Entropy Image (GEnI),Sagittal,{CASIA-B, SOTON}
E. Zhang et al. \nocite{zhang2010active},2010,TS,Active Energy Image (AEI),Sagittal,{CASIA-B, CASIA-C}
T. Lam et al. \nocite{lam2011gait},2011,TS,Gait Glow Image (GFI) using optical flow,Sagittal,USF v2.1
Zheng et al. \nocite{zheng2011robust},2011,TS,View transformation model using GEI,Multiview,CASIA-B
C. Wang et al. \nocite{wang2012human},2012,TS,Chrono-Gait Image (CGI),Sagittal,{USF, SOTON, CASIA-B}
N Liu et al.\nocite{liu2013robust},2013,TS,GEI-based Multiview Subspace Representation,Multiview,{CASIA-B, CMU-MoBo}
Dupuis et al. \nocite{dupuis2013feature},2013,TS,GEI subsetting using random forests,Multiview,CASIA-B
Hofmann et al. \nocite{tum_gaid_paper},2014,TK,Depth Grad. Hist. En. Img. (DGHEI) with audio,Sagittal,TUM-GAID
Aqmar et al. \nocite{aqmar2014gait},2014,TS,Gait Fluctuation Image (GFlucI),Sagittal,{OU-ISIR-D, OU-ISIR-LP}
P. Arora et al. \nocite{Arora2015},2015,TS,Gait Information Image (GII) features,Sagittal,{OU-ISIR, SOTON, CASIA-B}
X. Xing et al. \nocite{Xing2015},2015,TS,Canonical correlation analysis of GEI,Multiview,CASIA-B
P Yogarajah et al. \nocite{Yogarajah20153},2015,TS,Joint sparsity model and $l1$-norm`,Sagittal,CASIA-B
S.D. Choudhury et al \nocite{choudhury2015robust} ,2015,TS,View-invariant multi-scale  gait recognition,Multiview,CASIA-B
Zhao et al. \nocite{Zhao2015},2016,TS,Walking Path Image (WPI),Multiview,CASIA-B
I. Rida et al. \nocite{rida2016human},2016,TS,GEI segmentation using group lasso motion ,Multiview,CASIA-B
K. Shiraga et al. \nocite{shiraga2016geinet},2016,TS,GEINet -- CNN on GEI  ,Multiview,OU-ISIR-LP
C. Li et al. \nocite{li2017deepgait},2017,TS,DeepGait -- VGG-D on GEI with Joint Bayesian,Multiview,OU-ISIR-LP
Z. Wu et al. \nocite{wu2017comprehensive},2017,TS,{Ensemble of CNN, GEI and temporal features},Multiview,{OU-ISIR-LP, CASIA-B}
\end{filecontents*}
\csvreader[late after line=\\]{GR-T.csv}
        {Study=\study, Year=\y, Type=\typ, Technique=\tech, Plane=\pl, Dataset=\ds}
        {\thecsvrow. & \y & \study & \tech & \pl & \ds}\bottomrule
      \end{tabular}
    \end{center}

    \bigskip \emph{Note:} The terms `$i$' and `$s$' stands for instances and human subjects respectively for the set of sequences (seq.) in the dataset. 
  \end{minipage}
\end{sidewaystable}

  




L. Wang et al. proposed a unique method to recognise gait by analysing
the contours of the silhouettes \cite{wang2003automatic}. The CASIA
dataset A was used in this experiment. The shape of the contour of a
given silhouette sequence is converted to a template with the use of
Procrustes shape analysis. Different exemplars are created for each
viewpoint. They tried three types of nearest neighbour algorithms,
viz., NN, kNN, and ENN (NN with class exemplar), in which ENN provided
the best results for gait recognition.

Experimental results of Cuntoor et al. \cite{cuntoor2003combining}
suggests that Dynamic Time Warping (DTW) and Hidden Markov Model (HMM)
can be combined to produce a better gait recognition result. DTW was
used to align the motion of the arms and legs to normalise the phase
of gait while HMM is used to define the leg dynamics. 

The efforts of the University of South Florida
\cite{sarkar2005humanid} has brought forth a new revolution to gait
recognition. They compiled the USF Gait Challenge dataset which
consists of 1,870 gait sequences obtained from 122 human subjects over
5 types of variation. The dataset was categorised to 12 challenge
probe sets for experimentation and a gallery set for training. A
simple baseline algorithm was developed to facilitate users of the
dataset to compare the performance of gait algorithms effectively. The
algorithm involves the use of the Tanimoto similarity measure
(Eq~\ref{eq:tanimoto}) to guage the similarity between two
silhouettes. 

\begin{equation}
  \label{eq:tanimoto}
  Sim(p,q) = \cfrac{|p \cap q|}{|p \cup q|}
\end{equation}

Here, $p$ and $q$ are two binarized images where each image is
represented as an ordered set of pixel values. Their intersection
corresponds to the number of pixels that are the same in their
silhouettes. Their union provides the total number of pixel space
taken when both silhouettes superimpose. The correlation between the
silhouette similarities provides the measure of closeness used for the
recognition step. Further details of the USF Gait Challenge dataset is
provided in section~\ref{sec:gait-datasets}.

The most notable form of silhouette-based gait recognition techniques
uses the production of a Gait Energy Image (GEI) template
\cite{man2006individual} from the gait cycle. Technically, the GEI
shows how energy is dissipated spatially through the stages of the
gait cycle. It is so prevalent in literature such that silhouette
based methods that are published after its time (2006) can be
classified either as GEI-based or non-GEI-based.

Given a sequence of $N$ homogeneous gait silhouettes of pure black and
white colour definition of a single gait instance, $B_t(x,y)$, the GEI
can be expressed as

\begin{equation}
  \label{eq:GEI}
  G(x,y) = \cfrac{1}{N}\sum_{t=1}^{N}B_t(x,y)
\end{equation}

It is created by superimposing the pixels of silhouette sequence of a
given gait by summing the values and averaging them resulting in the
output as a grey-level image of proportional pixel intensities. The
term homogeneous in here refers to the constraint that all images
of the ordered gait sequence must be of the same dimensions. Hence the
final stage boils down to image comparison of the test GEI with the
GEIs in the gait database. An example of a GEI obtained from a
subject's gait from the CASIA-B dataset along the sagittal plane is as
shown in Figure~\ref{fig:GEIseq}. The real GEI as illustrated is not
used for the comparison. Instead, a synthetic template is produced
with an addition of noise and removal of the feet region to cope up
with the changes in walking surface, shoe type, and
clothing. The features are learned based on Multiple Discriminant
Analysis (MDA) after passing through Principle Component Analysis
(PCA) for dimensionality reduction. The Euclidean distances towards
the class centres with respect to the features provide the closeness
measure for recognition. Experimental results of Ju Han et
al. \cite{man2006individual} show recognition rates averaging around
55.64\% using the USF Gait Challenge v2.1 database.

In about the same time, T. H. Lam et al. were developing their version
of gait feature templates using motion silhouette contour templates
(MSCTs) and static silhouette templates (SSTs) \cite{Lam_T}. The
characteristics that capture motion are represented in MSCT while SST
depicts the static characteristics of the human gait. They did
seemingly perform well for both indoor and outdoor applications. But
soon after the article was published just after, they noticed that
their method was not as accurate as the GEI. Hence they developed a
whole new technique called the gait flow image (GFI) based on optical
flows for a better performing biometric system \cite{lam2011gait}. For
Rank 1, the CCR of SST and MSCT was 29.75\% and 34.00 respectively,
whereas GEI in their experimental observation gave a CCR of
39.08\% which was outperformed by the GFI method with a recognition
rate of 42.83\%.

X. Yang et al. \cite{Yang_X} show that the original GEI technique can
be enhanced if just the dynamic area of the GEI template is extracted
for feature production. Using this enhancement, they were able to
produce an effective recognition rate of 55.64\% tested against the
USF Gait Challenge v2.1 database. Hofmann et
al. \cite{hofmann2011gait} combine colour histograms with the GEI to
be able to identify and track gait even through the situation of
occlusion. This method was verified using their dataset called
TUM-IITKGP.

The many studies then published their own version of an energy image
adopted from the core principles of the GEI. One of such proposals
includes the GFlucI -- gait fluctuation image \cite{aqmar2014gait}. By
applying a technique called Self-DTW, a variant of DTW, the silhouette
sequences are phase-normalised. These sequences are then converted to
a template called the GFlucI which incorporates only the spatial and
temporal fluctuations in gait. Once a GEI template is formed, the
temporal characteristics of gait are lost. A chrono-gait image (CGI)
was proposed to incorporate this temporal information to the GEI in
\cite{wang2012human}. The motion history image (MHI)
\cite{ahad2012motion} originally proposed for action recognition is
specialised for gait recognition in \cite{Nguyen2014}. More examples
include the gait entropy image (GEnI) \cite{bashir2010gait} and active
energy image (AEI) \cite{zhang2010active}.

The information from a depth camera can be combined with that of an
RGB camera to provide a richer set of features for gait
recognition. Hofmann and his team \cite{hofmann20122} have applied
this notion and formulated a new technique for producing a GEI-like
image called the depth gradient histogram energy image (DGHEI). It
is produced by collating the histogram of oriented gradients (HOG) of
every image in the gait sequence to produce a single gait template for
comparison. Despite the fact of being a weaker biometric indicator,
audio-based data from gait, by itself, can estimate a range of soft
biometrics such as age, height, gender, and shoe type.  Later in
\cite{tum_gaid_paper}, DGHEI was combined with audio data to be able
to extract further distinguishing information to aid in gait
recognition.

There is an inherent uncertainty in certain regions associated
with a gait feature template for a given individual. This notion is
exploited by P. Arora et al. \cite{Arora2015}. Their work involves the
production of a GEI-like gait template over a single gait cycle called
the gait information image (GII) to generate features based on
Hanman–Anirban entropy function \cite{hanmandlu2011content} -- a
modified version of Shannon's logarithmic entropy function
\cite{shannon2001mathematical}. This GII gives rise to two feature
templates, the energy feature (GII-EF) and the sigmoid feature
(GII-SF). The statistics of the motion patterns of these two templates 
are hence used to make the prediction with the conventional nearest
neighbour classifier.

Re-identification is another important scenario in biometric gait
analysis. In a common video surveillance network, a person can be
observed from many angles as she/he moves from the visual range of one
camera to another. The challenge in addition to analysing the gait based
recognition is to be able to track the individual through the
different angles over the network of cameras. The work of Zheng Liu et
al. \cite{Liu20151144} addresses this research problem by combining
the GEI template's ability to match gait along with Gabor features and
HSV histograms to match appearance. Features are matched hierarchically
with descriptors for matching. The metric learning method was used as
an alternative to the nearest neighbour classifier.

\vspace{5pt}
\subsection{Non-template methods}
\label{sec:non-template}

Though found to be efficient in practice, not all silhouette-based
methods in literature involves the production of a template image. We
shall discuss some of them here.

J.P. Foster et al. \cite{Foster_JP} have claimed to have attained
recognition rates above 75\% by running experiments on 114 subjects of
the SOTON video gait dataset. Their method monitors the temporal
changes in the areas of the clipped gait window segmented by masked
sectors. Using these time-varying area metrics, they formulate a
feature vector for recognition.  

A framework for joint tracking and event detection using maximum a
posteriori (MAP) hypothesis was suggested in
\cite{wang2004framework}. Although not proposed as a gait recognition
technique, this method would be able to produce an efficient gait
recognition technique if utilised in such a way. Four events were
modelled corresponding to the different directions of walking
concerning the point of view: walking towards the camera, walking away
from the camera, walking from left to right and from right to
left. The corresponding HMM was trained and validated using the UMD
Dataset 1.

\begin{sidewaystable}
  \centering
  \renewcommand{\arraystretch}{1.1}
  \caption{Non-template Methods for Gait Recognition}
  \label{tab:non-template}
  \begin{minipage}{0.8\columnwidth}
    \footnotesize
    \begin{center}
      \begin{tabular}{rcllll}
        \toprule S.No. & Year & Study & Technique & Plane & Dataset \\
        \midrule
        \begin{filecontents*}{GR-N.csv}
Study,Year,Type,Technique,Plane,Dataset
J. Little \& J. Boyd \nocite{little1998},1998,NS,Phase difference of optical flows,Sagittal,UCSD
Johnson \& Bobick \nocite{johnson2001multi},2001,NS,Static body parameters,Multiview,Georgia Tech
A. Kale et al. \nocite{kale2002framework},2002,NS,Frame to exemplar distance (FED),Sagittal,{CMU, UMD}
J. P. Foster et al. \nocite{Foster_JP} ,2003,NS,Temporal change in segmented areas,Sagittal,SOTON: $28s \times 4i = 112$
L. Wang et al. \nocite{wang2003silhouette},2003,NS,PCA-based eigenspace transformation,Multiview,CASIA-A
A. Kale et al. \nocite{kale2004identification},2004,NS,FED with outer contour width,Sagittal,{CMU, USF, UMD}
R. Wang et al. \nocite{wang2004framework},2004,NS,Joint-tracking via MAP hypothesis,Multiview,UMD
N. V. Boulgouris et al. \nocite{Nikolaos_VB} ,2006,NS,Linear time normalisation,Sagittal,USF
Zongyi Liu \& S. Sarkar \nocite{liu2006improved},2006,NS,Population HMM,Sagittal,{USF, CMU, UMD}
N. V. Boulgouris et al. \nocite{Boulgouris} ,2007,NS,Matching of body components,Sagittal,USF
Zongyi Liu \& S. Sarkar \nocite{Zongyi_L},2007,NS,Population EigenStance-HMM,Sagittal,{USF, CMU}
Choudhury \& Tjahjadi \nocite{Choudhury_SD},2013,NS,Shape and motion analysis of contours,Sagittal,USF
Nguyen et al. \nocite{Nguyen2014},2014,NS,HMM of motion history image (MHI),Sagittal,CASIA-A
Chin P. Lee et al. \nocite{Lee2015},2015,NS,Transient binary patterns (TBP),Sagittal,{CASIA-B, OU-ISIR-D, CMU}
Zeng \& Wang \nocite{zeng2016view},2016,NS,Periodic deformation of human gait shape,Multiview,{CASIA-B, CMU-MoBo}
Muramatsu et al.\nocite{muramatsu2016view},2016,NS,Frequency domain features of joint-subspace,Multiview,OU-ISIR-LP 
\end{filecontents*}
\csvreader[late after line=\\]{GR-N.csv}
        {Study=\study, Year=\y, Type=\typ, Technique=\tech, Plane=\pl, Dataset=\ds}
        {\thecsvrow. & \y & \study & \tech & \pl & \ds}\bottomrule
      \end{tabular}
    \end{center}

\bigskip\bigskip

  \caption{Model-based Methods for Gait Recognition}
  \label{tab:model-based}
    \footnotesize
    \begin{center}
      \begin{tabular}{rclllll}
        \toprule S.No. & Year & Study & Type & Technique & Plane & Dataset \\
        \midrule
        \begin{filecontents*}{GR-M.csv}
Study,Year,Type,Technique,Plane,Dataset
Rawesak T. \& Bobick \nocite{tanawongsuwan2001gait} ,2001,Marker,Time-normalized joint trajectories,Sagittal,Georgia Tech
Z. J. Geradts et al. \nocite{geradts2002use},2002,Marker,Multi-planar gait parameter extraction,Multiview,$11s \times 2i = 22$ seq.
Rawesak T. \& Bobick \nocite{tanawongsuwan2003perf} ,2003,Marker,Speed-invariant gait parameters,Sagittal,$15s \times 36i = 540$ seq.
R. Zhang et al. \nocite{Zhang_R},2007,Silhouette,Five-link biped human model,Sagittal,{USF, CMU}
M. Goffredo et al. \nocite{Goffredo2008},2008,Silhouette,Geometric marker-less joint estimation,Multiview,CASIA-B
P. Chattopadhyay et al. \nocite{Chattopadhyay20159},2015,Kinect,Combining depth and silhouette features,Frontal,$29s \times 8i = 232$ seq.
D. Kastaniotis et al. \nocite{Kastaniotis2015},2015,Kinect,{Euler angles, dissimilarity space mapping},Oblique,{UPCVGait}
\end{filecontents*}
\csvreader[late after line=\\]{GR-M.csv}
        {Study=\study, Year=\y, Type=\typ, Technique=\tech, Plane=\pl, Dataset=\ds}
        {\thecsvrow. & \y & \study & \typ & \tech & \pl & \ds} \bottomrule
      \end{tabular}
    \end{center}

    \bigskip 
    \emph{Note:} Model-based features can either be extracted through 3D cameras with the subject fitted with IR or bulb \textit{markers}, articulation points inferred from \textit{silhouettes}, or depth information from Microsoft \textit{Kinect}. 
  \end{minipage}
\end{sidewaystable}

In the work by N. V. Boulgouris et al. \cite{Nikolaos_VB}, a low
dimensional feature matrix is represented by accounting the average
distances from the centre of the silhouette. Each silhouette is
represented as a feature vector which is composed of a sequence of
angular transforms made on segmented angular slices
$\Delta\theta$. The dimension of each vector is
$K = 360/\Delta\theta$. The feature matrix is produced by collating
the vectors of each silhouette in the sequence of frames. Hence each
column would represent a frame, which each row would depict the change
in the angular transform for a given angular slice. In the USF gait
challenge dataset, a period in the gallery (reference) is not equal to
that of the probe sequence (test). Hence, a technique called linear
time normalisation was utilised to make the feature matrix of each
probe and gallery sequence comparable by compensating for the
difference in the sequence length. The same method is used in their
next paper \cite{Boulgouris} but with a different feature extraction
technique. In this work, the body as depicted in the silhouette is
segmented into components. The components are ranked with according to
their proportional relevance during the comparison operation.

The work of Zongyi Liu and Sudeep Sarkar \cite{Zongyi_L} promoted the
fusion of face and gait biometrics. A subset of the USF dataset
was used for this purpose. The exemplars, in this case, are not obtained
from a single template but a specified stances analogous to the
formally depicted gait cycle. These stance frames are modelled using a
population EigenStance-HMM; a method that is extended from their
previous technique, population HMM \cite{liu2006improved}. The entire
population of the training set is fed in the learning phase. A given
gait silhouette sequence can be segregated into these stance models by
K-Means clustering with Tanimoto distance as the distance measure. A
cyclic Bakis variant of HMM was modelled for the gait
recognition. This gait recognition algorithm was used in combination
of face recognition using Elastic Bunch Graph Matching (EBGM) based on
Gabor features to attain a much higher performance over the Gait
Challenge baseline algorithm when compared using the toughest sets of
the USF Gait Challenge dataset.

A gait recognition that analyses both shape and motion is proposed in
\cite{Choudhury_SD}. The gait period here is depicted as ten
phases. Spatiotemporal shape features are obtained from these phases
in the form of Fourier descriptors. The silhouette contour at each
step is segmented by fitting ellipses. The similarity is calculated
utilising Bhattacharyya distance between the ellipse parameters
taken as features. Dynamic time warping (DTW) is applied to compare
leading knee rotation with relevance to arm swing pattern over a gait
cycle. DTW is used so as to counteract the effects of walking speed,
clothing, shadows, and hair styles.

\section{Model-based techniques}
\label{sec:joint-traj-meth}

While basic spatiotemporal methods give cues as to how the body
position vary with accordance to time as a whole, it would be much
more accurate to do a spatiotemporal analysis on all articulation
points separately. The implementation of that sort will fall under the
category of joint trajectory or model-based methods. That is, the
trajectory of each joint is tracked live and analysed as individual
components; efforts are made to model the human structure accurately.

Bulb markers can be attached to certain points on the body considered
necessary for gait analysis such as ankles, knees, hands, elbows,
shoulders and torso. When observed from a camera with a low exposure,
only the bulb illumination can be perceived. This method facilitates
an easier and a more accurate analysis of gait through computer
vision. Rawesak T. \& Bobick \cite{tanawongsuwan2001gait} implemented
this by strapping 18 human subjects with 16 markers at appropriate
locations and projected their gait at a sagittal angle. The sensors
were able to reconstruct a mobile skeletal structure recovered from
the joints. From this data, they were able to assess the articulation
points over time. The variance in time was normalised by applying DTW,
and the recognition was based on the nearest neighbour algorithm to
produce a modest recognition rate of 73\%.

A study by Z. J. Geradts et al. \cite{geradts2002use} was conducted to
be able to extract gait-related parameters from all three planes --
frontal, transverse and sagittal -- from surveillance cameras. 11
human subjects participated in the experiment and involved the use of
11 bulb markers fitted to the necessary points to each subject. They
were able to observe various parameters from step length, cycle time
and cadence to joint angles and spatial positioning. After a simple
analysis of variance (ANOVA) on the gait parameters extracted, it
seemed like the foot angle exhibited the most variance and then the
time average hip joint angles followed by the step length. Hence these
are considered to be the best parameters to be used for
recognition. However, their research concluded that the gait analysis
cannot be used for identification at that time (the year 2002).

Rawesak T. \& Bobick later produced a study on the recognition of gait
in different speeds of walking \cite{tanawongsuwan2003perf}, but this
time, they resorted to a more comfortable reflective suit for the
articulation point signal extraction. The experimental is described in
\cite{tanawongsuwan2003study}. A 12-camera VICON MoCap system is used
for the 3D motion analysis. The result based on 15 human subjects
concluded that a positive linear correlation could be observed between
cadence and speed, and a negative exponential correlation could be
observed between stride time and speed.

The methods described so far require the use of complicated hardware
for better accuracy. They either require reflectors, bulb
markers, or magneto sensors to be fitted on to the points of interest
of the human subject to gather the point-light information during
his/her gait. Due to their nature, these methods are not
practically feasible for biometric application. It is to note that the
concept of being an `unobtrusive' means of biometrics is lost here.

Not all model-based techniques, however, are impractical for
application. Zhang et al. \cite{Zhang_R} show that it is possible to
extract a five-link biped human model from a two-dimensional video
feed to produce a model-based gait recognition system. The five links
extracted consist of both the lower and upper leg portions and the
upper body. The Metropolis-Hastings algorithm \cite{metroHastings} was
used for the feature extraction. The arm motion is neglected in this
study. The changes in angular displacement of these links in the
sagittal elevation angle (SEA) are analysed to using the HMM for the
gait recognition. The method was applied to both CMU MoBo dataset as
well as USF Gait Challenge video dataset.
 
An innovative yet simple method for locating the articulation points
of the lower limb joints was implemented by M. Goffredo et
al. \cite{Goffredo2008}. By making smart estimates on where to
initialise the points, the point-light data for the hip, both knees
and ankles were extracted. This method was able to extract these
points over multiple views as provided by the CASIA-B video
database. By recording the temporal changes of these points, a profile
recorded could be recognised with the help of the k-nearest neighbour
algorithm (kNN).

A standard video feed would provide a two-dimensional data for
processing. When added with depth based information, more accurate
conclusions can be drawn to aid recognition. Microsoft Kinect provides
this functionality. The Kinect is used in references
\cite{Chattopadhyay20159} and \cite{Kastaniotis2015} by facilitating
three-dimensional data flow for a more natural and efficient biometric
gait recognition.

\begin{table}
\footnotesize
\begin{minipage}{\columnwidth}
\begin{center}
  \caption{Datasets for Gait Analysis} 
  \renewcommand{\arraystretch}{1.2}
  \begin{tabularx}{1.0\linewidth}{
    l 
    c 
    c 
    r 
    r 
    r 
    l 
    l 
    } 
    \toprule
    Database & Year* & Env & View  & Sub. & Seq. & Other variations \\
    \midrule
    \begin{filecontents*}{Datasets.csv}
Database,Year*,Env,View,Covariates,Sub,Seq,Sequences
UCSD  \cite{little1998},1998,FO,1,,6,42,$6s \times 7i = 42$
MIT AI  \cite{lee2002gait},2001,FI,4,{Time},24,194,$24s \sim 194$
CMU MoBo  \cite{cmu_mobo},2001,TI,6,{Speed(2), load(2), inclination(2)},25,600,$25s \times 4i \times 6v = 600$
HID Georgia Tech  \cite{gtech},2001,FIO,3,{Time},20,$>$280,$20s \sim 280+$
HID-UMD Dataset 1  \cite{umd},2001,FO,4,,25,100,$25s \times 4v = 100$
HID-UMD Dataset 2  \cite{umd},2001,FO,2,Time(2),55,220,$55s \times 2v \times 2i = 220$
CASIA Dataset A  \cite{wang2003silhouette},2001,FO,3,,20,240,$20s \times 4i \times 3v = 240$
SOTON Small DB  \cite{hidgait},2002,FI,4,{Load(4), clothing(3), footwear(6), speed(3)},12,n/a,$12s$
SOTON Large DB  \cite{soton_paper},2002,FTIO,2,{Terrain(3)},114,2128,$114s \sim 2128$
USF Gait Challenge  \cite{sarkar2005humanid},2002,FO,2,{Terrain(2), footwear(2), load(2), time(2)},122,1870,$122s \sim 1870i$
CASIA Dataset B  \cite{yu2006framework},2005,FI,11,{Clothing(2), load(2)},124,13640,$124s \times 11v \times 10i = 13640$
CASIA Dataset C  \cite{tan2006efficient},2005,FO,1,{Speed(3), load(2)},153,612,$153s \times 4i = 612$
CASIA Dataset D  \cite{zheng2011evaluation},2009,FI,1,,88,880,$88s$
OU-ISIR Speed Transition  \cite{ouisir_st},2007,FTI,1,Speed shift,179,n/a,$179s$
OU-ISIR Large Population  \cite{ouisir_lp},2009,FI,2,,4007,7842,$4007s \sim 7842$
TUM-IITKGP  \cite{hofmann2011gait},2010,FI,2,{Load(2), occlusion},35,840,$35s \times 6c \times 2v \times 2i = 840$
OU-ISIR Inertial Sensor  \cite{ouisir_is},2011,FI,,Inclination(3),744,3468,$744s \sim 3468$
SOTON Temporal  \cite{matovski2012effect},2011,FI,12,{Time(5), clothing(2), footwear(2)},25,26160,$25s \sim 2180i \times 12v = 26160$
OU-ISIR Treadmill - A  \cite{ouisir_tm},2012,TI,1,Speed(9),34,612,$34s \times 9v \times 2i = 612$
OU-ISIR Treadmill - B  \cite{ouisir_tm},2012,TI,1,Clothing(32),68,2176,$68s \times 32c = 2176$
OU-ISIR Treadmill - C  \cite{ouisir_tm},2012,TI,25, ,200,5000,$200s \times 25v = 5000$
OU-ISIR Treadmill - D  \cite{ouisir_tm},2012,TI,1,Gait fluctuations,185,370,$185s \sim 370$
TUM-GAID  \cite{tum_gaid_paper},2012,Fi,1,{Load(2), footwear(2), time(2), clothing(2)},305,3370,$305s \sim 3370$
\end{filecontents*}
\csvreader[late after line=\\]{Datasets.csv}
    {Database=\nam, Year*=\yr, Env=\ev, View=\vw, Covariates=\cvar,
    Sub=\sub, Seq=\seq}
    {\nam & \yr & \ev & \vw  & \sub & \seq & \cvar}
    \bottomrule                                 
  \end{tabularx}
  \label{tab:dataset}
\end{center}

\bigskip
  * The year indicates the year of recent release, not the year of its
  initial release. SOTON released its first version in 1996 and its
  first temporal set in 2003. OU-ISIR Treadmill datasets were
  iteratively developed from 2007-12. 

  \emph{Env Legends:} F - static flooring, T - treadmill, I - indoor, O -
  outdoor.
\end{minipage}
\end{table}

\section{Gait Datasets}
\label{sec:gait-datasets}

With the outbreak of a substantial amount of algorithms to analyse
gait there also comes a need to compare them. A standard gait dataset
would be able to serve means so as to benchmark such algorithms
especially in the case of biometric application. The categories of
datasets cover a wide range based on the needs of the gait analyst
from lightweight datasets to large-scale databases. An overview of
the gait databases described here is given in
Table~\ref{tab:dataset}. 

When DARPA launched the HumanID programme in the year 2000, many
institutions joined and released their first version of the database
in the year 2001. Institutions that released their dataset for public
usage include MIT, CMU, SOTON, Georgia Tech, UMD, USF, and
CASIA. Nearly all of the institutions whose databases are described in
this section are associated with this programme. The programme ended
in 2004 due to privacy issues \cite{hidgait}, but the databases
compiled as a result is still publicly available from the institutions
that developed them.

\subsection{First Datasets on Gait}
\label{sec:first-datasets}

When gait analysis was increasing in interest as a biometric, the
University of Southampton (SOTON) generated its gait dataset in the
year 1996 \cite{cunado1997using}. It had only four subjects with a
total of just 16 gait sequences. Self-occlusion occurs when the visual
perception of one leg overlaps with that of the other leg. The first
SOTON dataset attempted to reduce the self-occlusion problem by making
the subjects wear white tracksuits with a single dark stripe
vertically down either side walking frontoparallel with a plain cloth
background. The visible line can help differentiate and track the leg
closest to the camera throughout the gait cycle.

The first known dataset to become publicly available was developed by
the Visual Computing Group in the University of California San Diego
(UCSD) \cite{hidgait} in the year 1998. Its last update included six
subjects and seven instances each adding up to a total of 42
frontoparallel sequences (sagittal observation) \cite{little1998}. The
number of subjects is still considerably small for a provable
biometric application (for what it was intended). But the demands were
met at its time, and its response paved way for other gait datasets to
emerge.

MIT compiled a gait dataset with 24 persons at various trials
taken at four different views in its Artificial Intelligence
Laboratory \cite{lee2002gait}. With 194 gait sequences, it
became one of the first multi-view gait datasets to be used in its
time. Due to the arrival of many other large-scale datasets in about
the same time, the usage of the MIT gait database became diminished in
literature.

\subsection{Carnegie Mellon University (CMU)}
\label{sec:cmu}

The CMU Motion of Body (MoBo) database was compiled in the Robotics
Institute, Carnegie Mellon University. It is a treadmill based
dataset, i.e., the gait sequences are recorded with the subjects
walking on the treadmill. One gait instance, $i$, was taken for each
of four gait variations recorded for 25 subjects, $s$ taken at six
viewpoints, $v$. \[ 25s \times 4i \times 6v = 600 \text{ seq.}\] The
variations are slow walk, fast walk, incline walk, and walk with the
ball \cite{cmu_mobo}. The videos are recorded inside the lab (called
the CMU 3D room). The CMU MoBo database was one of the first
public databases released as part of the HumanID programme it easily
became the first to be accessed by a wider range of people.

\subsection{Georgia Institute of  Technology}
\label{sec:georgia-tech}

The dataset released by Georgia Tech, HID (Human
ID), contains both video and motion capture data. There were a total
of 20 subjects but were split in different ways according to the group
of gait data collected. 15 members took part in the outdoor and
outdoor video gait database accounting up to 268 gait instances. 18
members wore a magnetic sensor suit for motion capture compiling 20
gait instances wherein 2 of the subjects have additional motion capture
data taken at different times. There are also bonus videos recorded as
part of calibration which could also be used for gait analysis. 

The outdoor data is collected from the terrace of the building while
the indoor data is taken in lab conditions in front of a white wall
and grey carpet. MoCap videos were recorded in a green room while the data
itself is available in MAYA format. More detailed information on
database description can be found in \cite{gtech}.

\subsection{University of Maryland (UMD)}
\label{sec:hid-umd}

The gait database designed by the University of Maryland (UMD) has two
datasets; the first one with multiple views and the second one with
more subjects \cite{umd}. 

\subsubsection{Dataset 1}

The gait of 25 subjects, $s$, were observed in 4 different directions,
$v$: frontal anterior, frontal posterior, frontoparallel left and
frontoparallel right. The gait sequences were recorded outdoors in an
environment that seems to resemble a courtyard. Thus the total
sequence count becomes 
\[ 25s \times 4v = 100 \text{ seq.}\]

\subsubsection{Dataset 2}

In an environment similar to that of the previous dataset, the gait of
55 subjects, $s$ were recorded in two directions, $v$, while walking along a
T-shaped pathway. The two views were orthogonal to each other and
captured simultaneously. The dataset accounted two gait instances $i$ for
each subject which brings the sequence count to be 
\[55s \times 2i \times 2v = 220 \text{ seq.} \]

\subsection{University of Southampton (SOTON)}
\label{sec:soton}

The HumanID programme caused SOTON to reformulate its dataset with
more individuals and reduced clothing constraints. With 28 subjects
and four sequences per subject, the size of the dataset came up to 112
sequences in the year 2001. This database was used in \cite{Foster_JP,
  Hayfron_A, sarkar2005humanid}. SOTON again revamped its database in
the year 2002. The gait database was structured in two different
datasets called the SOTON small and SOTON large.

\subsubsection{Large Dataset}

The larger dataset was designed with a combination of indoor and
outdoor scenarios of 114 subjects \cite{soton_paper}. The following
are the different walking conditions covered:

\vspace{8pt}

\renewcommand{\arraystretch}{1.2}
\begin{tabularx}{0.9\linewidth}{ll>{\raggedright\arraybackslash}X}
  \tabitem terrain &:& indoor flooring, indoor treadmill, outdoor asphalt \\
  \tabitem direction &:& towards left of camera, \newline towards right of
                camera \\
  \tabitem viewpoint &:& normal, oblique \\
\end{tabularx}

\vspace{8pt}

The indoor background is as same for the small database, but the
outdoor background has vehicles and pedestrians moving across at a
distance behind the subject being observed. The level of dynamic
activity in the outdoor variation is intended to pose a natural
challenge for computerised gait recognition. The total size of the
dataset exceeds 350 GB. References \cite{tum_gaid_paper,
  makihara2015gait} report a total of 2128 gait sequences in this
dataset.

\subsubsection{Small Dataset}

Though the large database had the most members, the small database
with 12 subjects had the most covariates compared to all other
currently available public databases. Gait sequences were recorded in
different conditions as follows:

\vspace{8pt}

\renewcommand{\arraystretch}{1.2}
\begin{tabularx}{0.9\linewidth}{ll>{\raggedright\arraybackslash}X}
  \tabitem load &:& handbag, barrel bag (in two poses), rucksack, no load\\
  \tabitem clothing &:& raincoat, trenchcoat, casual\\
  \tabitem footwear &:& flip-flops, bare feet, socks, boots, trainers, own shoes\\
  \tabitem speed &:& slow, quick, normal speed\\
  \tabitem direction &:& towards left of camera, \newline towards right of
                camera \\
  \tabitem viewpoint &:& normal oblique, normal elevated, frontal view \\
\end{tabularx}

\vspace{8pt}

The background is staged with a green curtain and a dark green
walkway. There is not yet a credible source to state exactly how many
gait sequences are present in this database. It is also to note that
not all 12 subjects perform all of the above variations; studies shows
records of only 11 subjects \cite{Arora2015, bashir2010gait}.

\subsubsection{Temporal Dataset}

\begin{table}
  \centering
  \caption{SOTON Temporal Gait Sessions}
  \begin{tabular}{|c||c|c|c|c|c|}
    \hline
    Session & Aug. & Sep. & Dec. & May & Aug \\
    \hline
    $t$ & 0 & 1 & 4 & 9 & 12 \\
    \hline
    \# Sub. & 25 & 23 & 22 & 21 & 18 \\ 
    \hline
  \end{tabular}
  \label{tab:sotontemporal}
\end{table}

The database consists of gait sequences from 25 subjects spanning over
a period of a complete year taken at 5 (irregular) intervals. 12
cameras ($v$) capture the gait of each subject. The sessions were
conducted as shown in Table\ref{tab:sotontemporal}. 20 instances, $i$,
were recorded for every subject for each session. Calculating with the
above information, we obtain the total number of sequences
\[(25s + 23s + 22s + 21s + 18s)\times 20i \times 12v  = 26160\text{ seq.}\]

The observation was done in the Multi-Biometric Tunnel of the
University of Southampton \cite{sama2011acq}. The background is tiled
with red, green and blue squares. This facilitates feature mapping to
consolidate the camera views to construct a 3-D image of anything that
passes through the tunnel. There were also slight specific variations
in clothing and footwear. Details of which are provided in
\cite{matovski2012effect}. This gait database is clearly the largest
to cover the temporal aspect of gait to date.

\begin{figure*}
  \centering
  \includegraphics[width=0.99\linewidth]{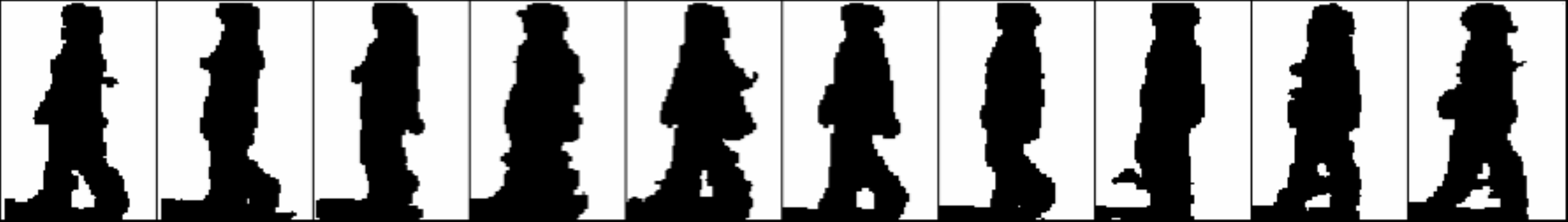}
  \caption{A sample sub-sequence of the silhouettes provided in the
    USF Gait Challenge database. This illustration contains one frame
    for every three frames in the sequence. Notice that the shadow
    produced by the subject is also captured leaving the heel angle
    incalculable for most of the sequences.}
  \label{fig:usfsample}
\end{figure*}

\subsection{University of South Florida (USF)}
\label{sec:usf-humanid}

The HumanID Gait Challenge dataset was compiled by the Computer
Vision and Pattern Recognition Group, University of South Florida
(USF). It was the first dataset to become immediately popular after the
start of the DARPA's HumanID programme. 122 subjects had their gait
sequences recorded in the following different conditions:

\vspace{8pt}

\renewcommand{\arraystretch}{1.2}
\begin{tabularx}{0.9\linewidth}{ll>{\raggedright\arraybackslash}X}
  \tabitem shoe type &:& A, B \\
  \tabitem load &:& with suitcase, without suitcase \\
  \tabitem terrain &:& grass, concrete \\
  \tabitem time &:& May 2001, Nov 2001 \\
  \tabitem viewpoint &:& sagittal left, sagittal right \\  
\end{tabularx}

\vspace{8pt}

All of the above come together to make 32 possible combinations, but
not all of them are applied to each of the subjects resulting in 1870
sequences. The bulk of video data adds up to 1.2 terabytes
\cite{usfgaitbaseline}. One must follow a complicated and
time-consuming process to get access to this video dataset, but USF
provides free access to the silhouettes of the videos processed using
the Gait Baseline algorithm \cite{sarkar2005humanid}. A sample
sub-sequence of the silhouettes is shown in Figure~\ref{fig:usfsample}
This silhouette dataset is what is used by most of the published
literature that utilises the USF Gait Challenge dataset
\cite{kale2004identification, liu2006improved, Nikolaos_VB,
  Boulgouris, Lam_T, Yang_X, lam2011gait, wang2012human}. Even the
famous GEI algorithm \cite{man2006individual} used not the video, but
only the silhouette database of the USF.

The foot angle was found to have the most variance among other
primitive gait measures \cite{geradts2002use}. This fact is astounding
since the foot angle is a highly sensitive feature that would be
difficult to be measured with an acceptable accuracy through the
silhouette based methods. The extraction of this feature would become
nearly impossible with the silhouette data readily available as part
of the USF Gait Challenge dataset. This is because most of the
silhouettes in this dataset is produced in such a way have the foot
occluded with shadows in which the boundary between the two is not
well defined.

\subsection{ Institute of Automation, Chinese Academy of Sciences  (CASIA)}
\label{sec:casia}

CASIA has developed four public gait databases till date. The
development started in 2001 at the National Laboratory of Pattern
Recognition (NLPR). In addition to the video, the silhouettes for each
of the following datasets are freely available for download.

\subsubsection{Dataset A}

This dataset was previously known as the NLPR Gait Database, early
literature such as \cite{wang2003silhouette} and
\cite{tan2006efficient} use this name. CASIA-A is a basic gait dataset
with the gait of 20 subjects, $s$, recorded at 3 views, $v$. 4
instances, $i$, were recorded for each subject leading to sequence
count of 
\[20s \times 4i \times 3v = 240 \text{ seq.}\]

The three views -- sagittal, 45$\degree$ and 90$\degree$ -- are captured
simultaneously in an outdoor environment (in front of a building).

\subsubsection{Dataset B}

\begin{figure*}
  \centering
  \includegraphics[width=0.99\linewidth]{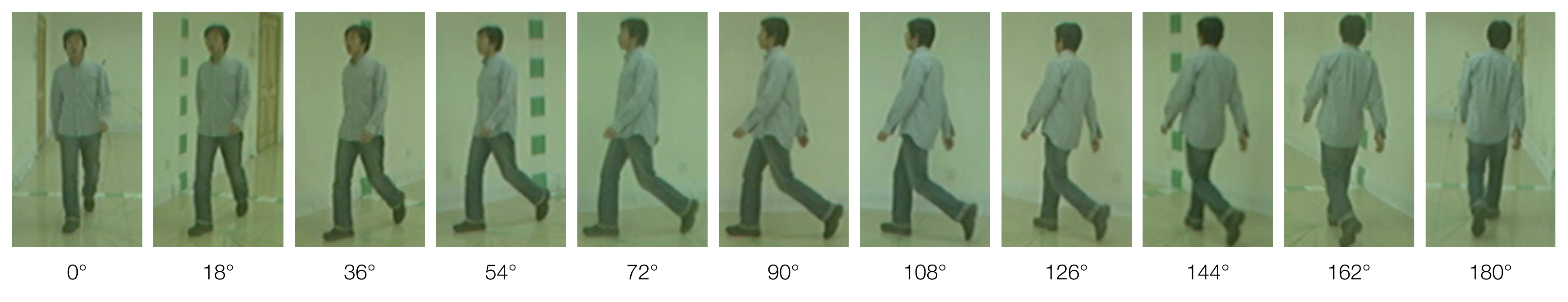}
  \caption{Cropped frames from CASIA Dataset B. A frame is taken from
    each of the 11 angles for a single individual at the same time.}
  \label{fig:casia-b}
\end{figure*}

With 124 individuals ($s$) each performing 10 gait instances ($i$)
observed in 11 simultaneous angles, the total number of sequences for
the CASIA-B dataset would be
\[ 124s \times 10i \times 11v = 13640 \text{ seq.}\]

The ten instances can be split up as

\begin{itemize}
\item 6 normal instances
\item 2 instances wearing a coat
\item 2 instances carrying a bag
\end{itemize}

Along with an additional video of bare background, the total number of
videos actually becomes 15004 (approximately 17.4 GB). The background
is mostly plain light green with two stage markings covering both the
wall and the floor \cite{yu2006framework}.  This is the most common
multiview database in literature. Figure~\ref{fig:casia-b} shows a
depiction of the dataset for a given time instant from each of the 11
angles. 

\subsubsection{Dataset C}

The gait sequences of 153 subjects, $s$, were collected. Each subject
performed 4 gait instances in different conditions, $c$: slow walk,
normal walk, fast walk, and walk while carrying a bag
\cite{tan2006efficient}. Unlike CASIA-A and CASIA-B, this dataset is
only observed at a single sagittal angle. Hence the total number of
gait sequences becomes
\[ 153s \times 4c = 612 \text{ seq.}\]

The sequences were recorded outdoors at night using an infrared
camera.

\subsubsection{Dataset D}

This dataset of synchronised footprint scans and gait images of 88
subjects. The camera observed the user from a sagittal angle while the
Rscan Footscan scans the foot placement with an equal sampling rate
while the subject progresses through the gait cycle. 10 trials are
taken per subject on average resulting in a total of around 880
sequences. CASIA also hosts a MATLAB mat file that contains the data
and can be freely downloaded from its site.

\subsection{Osaka University -- Institute of Scientific and Industrial Research
(OU-ISIR)}
\label{sec:ou-isir}

The OU-ISIR produced a whole range of biometric
databases. The ones that directly correspond to gait analysis are
provided in this section. The treadmill datasets were prepared over a
long course of time (from 2007 to 2012). The information for all
treadmill datasets can be obtained from \cite{ouisir_tm}. Only the
size-normalised silhouettes will be publicly available, not the videos
themselves, leaving the background irrelevant for our discussion.

\subsubsection{Treadmill dataset A -- Speed Variation}

34 subjects, $s$, walk on a treadmill with the speed periodically
increasing from 2 km/h to 10 km/h at intervals of 1 km/h. Hence 9
different speed variations, $c$, are accounted at a sagittal
angle. This process is repeated twice having two instances, $i$, of
each variation. This results in the number of gait sequences as 
\[ 34s \times 9c \times 2i = 612 \text{ seq.}\]

\subsubsection{Treadmill dataset B -- Clothing Variation}

This consists of 68 subjects, $s$,  with 32 combinations of clothing
styles, $c$, recorded along a sagittal angle resulting in a total
sequence count of
\[ 68s \times 32c = 2176 \text{ seq.} \]
 
The various clothing styles are as listed in
Table~\ref{tab:ouisir-b}.

\begin{table}
  \centering
  \caption{Clothing styles in OU-ISIR-B Dataset \cite{ouisir_tm}}
  \begin{tabular}{l|l|l}
    \hline
    RP: Regular Pants &  BP: Baggy Pants & SP: Short Pants \\
    Sk: Skirt & CP: Casual Pants &  HS: Half Shirt \\
    FS: Full Shirt & LC: Long Coat & Pk: Parker \\
    DJ : Down Jacket & CW: Casual Wear & RC: Rain Coat \\
    Ht: Hat & Cs: Casquette Cap & Mf: Muffler \\
    \hline
  \end{tabular}
  \label{tab:ouisir-b}
\end{table}

\subsubsection{Treadmill dataset C -- View Variation}

200 subjects are recorded in 25 different points of view synchronously
accounting for 5000 gait sequences in total. The dataset also covers a
substantial variation of age group from 4 to 75 years old. Male and
female subjects are equally distributed. This dataset would hence
facilitate age group and gender detection (soft biometric) rather than
gait recognition as only one instance is supplied per subject. 

Apart from its description, the OU-ISIR treadmill dataset C is still
not released publicly. On the time of writing this paper, the dataset
is still under preparation. 

\subsubsection{Treadmill dataset D -- Gait fluctuations}

The features of a gait cycle, in general, would not be the same in
every period of an individual's gait. A certain amount of fluctuation
can be observed. This dataset aims to measure this fluctuation with the
aim to use this feature as a biometric attribute. 185 subjects, $s$,
were recorded at the sagittal angle to measure fluctuations that could
occur in multiple gait periods. Each member had two trials, $i$
resulting in a number of sequences of
\[ 185s \times 2i = 370\text{ seq.}\]

Gait fluctuations are measured in terms of Normalised AutoCorrelation
(NAC) \cite{ouisir_tm}. The subjects were grouped into that with
higher NAC (least fluctuating), DB\textsubscript{high} and that of
lower NAC (most fluctuating), DB\textsubscript{low}. There were 100
members in each group, and 4 trials were made for each, but 15 subjects
were part of both groups.

\subsubsection{Large Population Dataset}

This dataset consists of gait silhouettes of over 4000 subjects with
ages of 1 to 94 years. The gait sequences were captured with two
cameras at two different angles out of which only one is available so
far as part of the publicly available dataset. This is the gait
dataset that involves the largest number of human subjects till date.
The total number of gait sequences in this dataset is reported to be
7842 \cite{makihara2015gait, tum_gaid_paper}. A detailed description
of this dataset can be found in \cite{ouisir_lp}.

\subsubsection{Speed Transition Dataset}

While all other datasets concentrated on keeping the speed relatively
constant for a given gait sequence, this dataset aims to observe the
gait while allowing changes in speed within a single gait
sequence. This is separated into of two datasets. In dataset 1, 179
subjects are made to walk towards a wall from a given point while
exhibiting gradual decrease in walking speed. While dataset 1 was
recorded indoors on flat ground, dataset 2 was
treadmill-based. Dataset 2 consisted of gait sequences from 178
subjects. Each gait sequence included either of acceleration or
deceleration of walking speed between 1 km/h to 5 km/h. Both datasets
have a gallery set which includes the subjects walking at a constant
speed of 4 km/h \cite{ouisir_st}.

There aren't much reported research that concerns the use of this
particular dataset which could be due to the level of challenge to be
faced by algorithms to recognise gait in such constraints.  The rare
use of this database could also be due to the extent of artificial
constraints applied to the subjects. The subjects have to control
their gait effectively to be conscious about the speed
transition. This anticipation could inhibit their natural gait.

\subsubsection{Inertial Sensor Dataset}

Unlike the other datasets, this one includes the kinetic data from
gait. The apparatus is in the form of a belt that has three IMUZ
sensors and a smartphone attached. Each IMUZ sensor consists of a
triaxial accelerometer and a gyroscope. The subject wears the belt
around the waist as he/she walks through the specified track. It is
claimed to be the largest ever inertial sensor-based gait dataset
\cite{ouisir_is}. The dataset is split into two subsets. The first
subset has 744 subjects with two trials of flat-ground gait. The
second subset is of 495 subjects two trials for no inclination, one
for up-slope walk and another for down-slope walk. Thus an estimate of
the total number of gait sequences would come across
\[ (744s \times 2i)+(495s \times 4i) = 3468\text{ seq.}\]

\subsection{Technical University of Munich (TUM)}
\label{sec:tum}

\subsubsection{TUM-IITKGP}

\begin{figure*}
  \centering
  \includegraphics[width=0.7\linewidth]{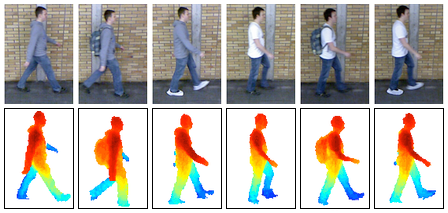}
  \caption{A sample from the TUM-GAID database with a frame taken from
    each of the 3 conditions taken at both January session (first
    three) and April session (last three). The bottom set of images
    correspond to the respective depth image at the same instant.}
  \label{fig:tumgaid}
\end{figure*}

The joint research efforts of the TUM and the Indian Institute of
Technology Kharagpur (IIT-KGP) has brought forth a gait database which
has become the first to address the problem of occlusion in gait
recognition \cite{hofmann2011gait}. Each of the 35 subjects, $s$,
followed 6 different walking conditions, $c$, two trials each for left
to right and right to left directions. Hence 4 instances are recorded
for each of the configuration per subject. The total number of gait
sequences becomes
\[35s \times 6c \times 4i = 840 \text{ seq.}\]

The configurations are regular, hands in pocket, wearing
a backpack, wearing a gown, dynamic occlusion, and static occlusion.
Dynamic occlusion refers to the scenario when the subject under
observation is occluded by a moving object. In this case, dynamic
object relates to another individual moving in the opposite direction
and parallel to the subject. Static occlusion is the scenario where
the subject is temporarily occluded by an object that doesn't
move. The static object in here refers to another individual standing
in front of the camera so as to block a portion of the gait sequence
capture.

\subsubsection{TUM-GAID}

This is a database that was created with an intention to recognise
gait from audio, image and depth (GAID) information
(Figure~\ref{fig:tumgaid}). The gait data was compiled from 305
subjects in three walking conditions (normal, backpack, coating
shoes). While the image and depth data have been studied before, the
audio data would facilitate acoustic gait recognition
\cite{tum_gaid_paper}. The variation in the dataset is as given below.

\vspace{8pt}

\renewcommand{\arraystretch}{1.2}
\begin{tabularx}{0.9\linewidth}{ll>{\raggedright\arraybackslash}X}
  \tabitem load &:&  with backpack ($\sim 5$ kg), \newline without backpack \\
  \tabitem footwear &:& regular shoes, coating shoes \\
  \tabitem clothing &:& winter attire, casual attire \\
  \tabitem time &:& Session 1 (January 2012),\newline Session 2 (April 2012) \\
\end{tabularx}

\vspace{8pt}

176 subjects were recorded in the January session (at -15\degree C)
and 161 subjects in the April session (at 15\degree C). 

There were five trials in each subject per session. Each trial consists
of walking once towards the right of the camera and another towards
the left. Three trials were recorded for normal walk, one trial while
wearing coating shoes over their own shoes, and one trial while
wearing a backpack. The total number of sequences can be calculated as
\[(176s + 161s) \times 5c \times 2i = 3370\text{ seq.}\]

The different clothing conditions are also associated with the weather
of the sessions. Subjects wore heavier winter clothing in the first
session while a more casual clothing in the second session. 32 of the
subjects were common in both sessions. Hence the data from these
members can be used for analysing the variations in gait with respect
to the differences in clothing and time.

\subsection{Other Gait Datasets}
\label{sec:other-datasets}

Recent publications show proposals of datasets that are not as
prevalent in use as the ones illustrated in
Table~\ref{tab:dataset}.

Jordan Frank compiled an accelerometer-based gait dataset in McGill
University in July 2010 \cite{frank2010gaitdata}. It is freely
available for download without the need of a license. The data was
obtained from mobile phones carried by the subjects and is extracted
using an open-source Android app called HumanSense. The dataset
contains the triaxial accelerometer readings for 20 subjects. Each
gait sequence is a 15-minute walk. Gait sequences of each individual
were captured in two sessions taken at different days. The dataset
also includes the gender, height, weight, clothing and shoe
descriptions of each subject. Frank show how this data can be used for
both activity recognition and gait recognition in
\cite{frank2011activity}.

The Multimedia University, Malaysia (MMU) developed a gait database
known as MMUGait DB \cite{ng2014development}. The database can be
split into two sets, MMUGait Large DB with 82 subjects, and MMUGait Gait
Covariate DB with 19 subjects. Each instance is recorded from the side
view and oblique view. The MMUGait Large DB contains at least 20 gait
sequences per subject in both towards left and right directions. The
MMUGait Covariate DB consist of 10 sequences for each of the 11
covariates. These include three types of clothing, four types of load,
two walking speeds, and two types of footwear. Their database,
however, is not publicly accessible.

A team lead by Kastaniotis in the University of Patras Computer Vision
Group (UPCV) compiled a Kinect-based gait dataset called UPCVgait. As
the data is recorded with a Microsoft Kinect and it includes both
depth information as well as RGB videos. 5 gait sequences were
captured for each of the 30 subjects. It was first used in
\cite{kastaniotis2013gait} for gender and pose estimation. It was also
used in \cite{Kastaniotis2015} for gait recognition. This dataset is only available at
\cite{upcvgait}.

\section{Discussion}
\label{sec:discussion}

Ever since the time of its introduction, gait analysis was largely of medical relevance. It was only at the break of the 21st century did gait analysis also become of increasing interest outside the clinic and into forensic science. The human body is designed to walk in a way such that the energy that is expended is kept to a minimum. The dissipation of energy is dependent on various factors such as the biological structure, load exerted, and the experience of the individual. These differences make the natural gait of each person to be distinct. This distinction is what analysts try to capture as gait biometrics.

Biometric gait recognition is yet to reach its peak accuracy. The state of the art published literature shows groundbreaking results. Nevertheless, they have been produced with datasets that are imposed by serious constraints. Even with the complete description of the USF Gait Challenge, papers show conflicting results with the same set of algorithms. A thorough, unbiased, practical evaluation of the recent methods is much of need at this time. This assessment should also cover validation through more than a single dataset.  It is also questionable whether the true natural gait of each subject is captured in the popular gait datasets. During the start of the walk, an individual would have adequate control over his/her gait. The currently available datasets do not always take this issue into account. Though datasets have attempted to account for the inhibiting factors of gait, there is still the problem of internal and external control that one can exhibit in his/her gait.

There was a study which took place at the end of the 20\textsuperscript{th} century which compares the performance of 3D camera systems available at that time \cite{ehara1997}. However, the technology has evolved and many new techniques have arrived today leaving the results obsolete. Hence a similar study remains yet to be compared to evaluate the performance of the state of the art MoCap systems.

\section{Research Directions}
\label{sec:research-dir}

Gait analysis is an interesting study and is proven to be applicable
for diverse domains of applications. There are many open areas of
research from which one can choose in the field. We have segregated them into two
groups -- active areas and those that are least explored.

\subsection{Active areas}
\begin{itemize}
\itemsep4pt

\item \textit{Reliable recognition independent of clothing style.}
  Several algorithms have been outlined to combat this problem
  (e.g., \cite{yu2006framework,Yogarajah20153}) and a handful of
  datasets to help with this regard (CASIA-B, OU-ISIR-B, TUM-GAID,
  and so on). However, state of the art gait biometrics algorithms do suffer a
  significant depreciation when the clothing style changes.


\item \textit{Gait recognition at different walking speed.}  With
  datasets like CASIA-C and OU-ISIR-A, gait recognition at different
  speeds is becoming more of interest
  \cite{tanawongsuwan2003study,iwashita2015gait,tanawongsuwan2003perf}. There
  are, however, further improvements that can be made.


\item \textit{View-independent gait recognition.} Numerous studies
  that propose recognition models that could cope with multiple
  views. Many of which can be observed in
  Tables~\ref{tab:template-based}, \ref{tab:non-template} and \ref{tab:model-based}. Steady progress is still being made in
  this area.

\item \textit{A comparative study between model-based and model-free
    recognition methods.} Though studies contrast on the merits and
  demerits of both approaches, there is still space for an in-depth
  comparative analysis with newer technology such as the upgraded versions of Kinect.

\item \textit{Unbiased comparison of state of the art gait
    recognition.} Many of the studies discussed in this paper report
  conflicting results of correct classification rates when
  performances of established algorithms are compared, even with the
  same dataset. The existing algorithms are to be compared with
  multiple datasets and without bias.

\item \textit{Gait reidentification performance.} Gait identification with occlusion can be a challenge by itself wherein multiple cycles are provided for training per person. However, identifying whether two given gait signatures are from the same person can be even more difficult especially when considering multiple views. 

\end{itemize}
 
\subsection{Least explored}
\begin{itemize}
\itemsep4pt


\item \textit{Characteristic differences in gait based on ethnicity.}
  People from different regions do have a characteristic gait. CASIA
  datasets are composed of Chinese individuals, while the USF
  gait dataset are largely American. More datasets are to be created
  across various ethnicity to study the associated differences between
  them in terms of their gait.

\item \textit{Ranking of factors that inhibit gait.} The confounding
  factors of gait are as listed in
  section~\ref{sec:inhibiting-factors}. Though known, they are not
  compared in a way that would determine to which extent they inhibit
  gait. For instance, to what instance could training affect biometric
  gait? This has been an open question to which no authoritative reports are
  currently available.

\item \textit{Different machine learning methods for gait
    recognition.} The novelty of the recognition algorithms lies on
  the technique in which features are extracted. Commonly used
  classifiers are ANN, kNN, and SVM. A detailed analysis on how
  different classifiers effect the prediction rate through state of
  the art gait feature extraction algorithms is yet to be reported.




\end{itemize}

\section{Conclusion and Future Work}
\label{sec:conclusion}

There is no doubt that gait analysis is still a growing field with a
whole range of applicable areas. Ever since its boom in the second
millennium, it has captured so much attention from different domains.
It is of keen interest in modern medicine, security agencies,
military organisations, sports industry, bionic prosthetics, and
social analytics. So much progress has been made through time, and it
will continue to do so in the future.

This survey covered numerous papers published in gait literature with a particular focus on gait biometrics. 
Our objective in the near future is to do an empirical study on all existing gait templates and address at least two of the unexplored research areas: comparison of various machine
learning classifiers on gait features and ranking the factors that inhibit gait.

\bibliographystyle{IEEEtran}
\bibliography{../GaitAnalysis}

\end{document}